\documentclass[letterpaper]{article}

\usepackage{natbib,alifeconf}  
\usepackage{tabularx,tabulary}
\usepackage{graphicx}
\usepackage{amsmath}
\usepackage{subcaption}
\usepackage{hyperref}
\usepackage{array}
\usepackage{float}
\usepackage{multirow}

\title{Emergence of Novelty in Evolutionary
Algorithms}

\author{David Herel$^{1,}{^{3}}$ , Dominika Zogatova$^{1}$, Matej Kripner$^{2,}$\thanks{Work done during internship at GoodAI Research.} \and Tomas Mikolov$^{1}$ \\
\mbox{}\\
$^1$ Czech Institute of Informatics, Robotics and Cybernetics, Czech Technical University in Prague \\
$^2$ Charles University in Prague \\
$^3$ Faculty of Electrical Engineering, Czech Technical University in Prague\\
david.herel@seznam.cz} 

\begin{document}
\maketitle

\begin{abstract}
One of the main problems of evolutionary algorithms is the convergence of the population to local minima. In this paper, we explore techniques that can avoid this problem by encouraging a diverse behavior of the agents through a shared reward system. The rewards are randomly distributed in the environment, and the agents are only rewarded for collecting them first. This leads to an emergence of a novel behavior of the agents. We introduce our approach to the maze problem and compare it to the previously proposed solution, denoted as Novelty Search \citep{b3}. We find that our solution leads to an improved performance while being significantly simpler. Building on that, we generalize the problem and apply our approach to a more advanced set of tasks, Atari Games, where we observe a similar performance quality with much less computational power needed.
\end{abstract}

\section{Introduction}
In nature, individuals of one population compete with each other for survival, and only the fittest ones can pass their genes on to the next generation. Similarly, in evolutionary algorithms, it is the agents who are rewarded for being the fittest. In order to realize the rewards, an objective function is needed to evaluate the individual solutions based on their proximity to the optimum \citep{b1} and to guide us through the search space to a valid solution. Even though evolutionary algorithms have been proven successful in numerous optimization tasks \citep{b1}, the deceptiveness of the environment is rather problematic. In reality, increasing fitness does not always reveal the best path and can mislead us from finding the global optimum \citep{b2}.\\
Aiming to solve this problem, many algorithms were developed \citep{mitigate1}\citep{mitigate2} that significantly reduce the deceptiveness. But the underlying problem remains – the objective function may still misdirect the search toward a dead end. Actually, finding a solution to this problem requires thinking counter-intuitively and abandoning the objective completely, as Novelty Search does \citep{b3}. We can think of the Novelty Search as a divergent search technique applied to an evolutionary computation, which also introduces a new reward system. Instead of rewarding agents based on their proximity to the goal, agents are rewarded for being different from others in the population. In comparison with the traditional objective-based evolutionary processes, Novelty Search algorithms have demonstrated their efficiency in many highly deceptive problems \citep{b3}.\\
In this paper, we propose our own search method called Sugar Search, which aims to reproduce or surpass Novelty Search results without explicitly defining novelty. Thus, the behavioral novelty will emerge as a by-product of the environment and objective function definition. Building on the qualities of the Novelty Search \citep{b3}, we reproduce this method on the maze problem, which is identical to the mazes presented in \citep{b3}, and use it for comparison with our Sugar Search technique. Following the proof of concept and given the competitive results, we also conduct experiments with different reward densities and a combined fitness-based and Sugar Search approach in the maze environments.\\
In the second part of this paper, the main objective is to present a more complex problem that our search technique can be applied to. For this purpose, we present Sugar Search in the Atari games environment. We compare the results of our method to the fitness-based approaches, which have shown good results on some Atari games \citep{ALE}, as well as some reinforcement algorithms like DQN \citep{dqn} and AC3 \citep{a3c}. Because of the challenges that arise with agent allocation in such a complex environment, we also generalize our approach to this problem.\\

\section{Related work}
There have been several approaches to solve the problem of local minima by encouraging novelty. One of the most well-known ones is the Novelty Search algorithm \citep{b3}. It ignores the objective and instead optimizes how unique the individual's behavior is. It greatly differs from the random search, because it provides higher rewards to a diverse behavior. There is also an archive, which stores the previously explored areas of the search space. The novelty score is then counted as the average distance from the $k$-nearest neighbors. However, the distance metric is problem-dependent, which could be problematic \citep{b3}. Another interesting idea, which promotes the novel behavior, is the Behavior characterization algorithm \citep{b9}. This algorithm maps each individual behavior to a vector space. The search is then driven towards diversity in a metric space of these behaviors. Another approach to this problem is Curiosity learning. It works with intrinsic rewards that promote exploration \citep{curiosity-learning}. Building on Novelty Search, Surprise Search was later introduced as another divergent search technique working with the notion of surprise. It achieves an efficiency comparable to the Novelty Search and is also able to find solutions more frequently \citep{surprise_search}. \\
The ability to generate a diverse set of high-performing solutions in evolutionary algorithms is crucial and has been addressed in many so-called Quality-Diversity algorithms, such as the MAP-Elites algorithm \citep{map} or NS with Local Competition \citep{local}. In MAP-Elites \citep{map}, the search space is discretized into unique regions, for each of which the best-performing solution is identified. The selection of the fittest individuals withing a population is restricted to a specific feature, the fittest individuals are kept in a multidimensional archive and only replaced if they are outperformed. This way a diverse group of well-performing individuals with different features can develop in the domain space in a single run. However, the problem of local optima within the elite may still occur over time and the scalability to high-dimensional spaces is also problematic due to the exponential growth of the regions with increasing dimensionality. This problem is overcome by CVT-MAP-Elites algorithm \citep{CVT}, which extends the MAP-Elites algorithm and proposes a solution to reduce the number of regions in high-dimensional spaces.\\
The problem of the deceptive local optima arises not just in evolutionary algorithms, but also in reinforcement learning itself. The reinforcement approaches optimize only for the reward (e.g. DQN \citep{dqn}, A3C \citep{a3c}) often fail to learn behaviors and strategies to overcome the local optima and solve the task at hand \citep{NsInAtari}. To solve this issue, Novelty Search and other approaches such as episodic curiosity \citep{deepmind} were explored. In \citep{NsInAtari} instead of measuring the final position of the agent and comparing it to other individuals of the current generation and the ones in the archive, the Levenshtein distance between the sets of actions performed by the agents was used to measure the extent of behavioral novelty. The approach of comparing the sequences of actions instead of the final positions removes the need to detect the exact position of an agent for each frame, and therefore makes it easier to generalize Novelty Search to any Atari game.\\

\section{Our approach: Sugar Search}

\begin{figure}[h!]
\includegraphics[scale=0.18]{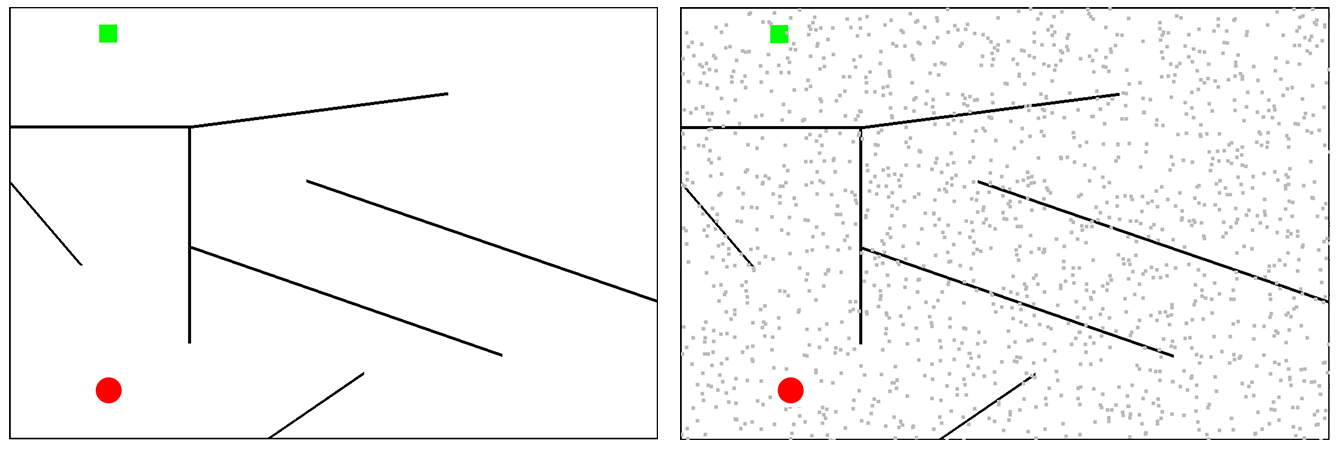}
\caption{Visualization of the hard map with (on the right) or without the sugars (on the left). The red dot is the starting point, the green dot is the goal, and the black lines are the obstacles. The sugars are represented as gray rectangles, they are placed randomly, and their density is the only hyperparameter. Only the agent who collects the sugar first is rewarded for it.}
\label{fig:sugars1}
\end{figure}

Even though Novelty Search is conceptually interesting, it seems to employ a certain degree of supervision, which is needed to measure and decide how unique each agent's behavior is. But this is inconsistent with what we observe in nature, where it is rather the environment that formulates, motivates, and rewards individuals for being different.\\
This difference can be demonstrated on a maze problem specified as follows. A maze contains multiple obstacles and a target. In the environment, there are agents whose goal is to reach the target while avoiding obstacles. If an agent reaches an obstacle, its movement stops.\\
In the Novelty Search algorithm, the agents that stop at a point far away from the others will receive a higher fitness, which motivates them to explore the maze. On the downside, this requires us to compute the $k$-nearest neighbors for each agent.\\
Our approach is fairly different. We propose a method called Sugar Search, which introduces sugar objects as a form of a reward. Sugars are distributed randomly (and uniformly) over the map, and an agent is rewarded only if he is the first one to collect a sugar object. This way the only form of supervision is the sugar placement and, compared to Novelty Search, we can avoid the costly computation of the $k$-nearest neighbors and eliminate the archive completely. The concept of the sugar-based search leads to a different pattern in the exploration of the search space than Novelty Search does.\\
Our fitness function $p$ for an agent $x$ is defined by:
\[p(x)=\sum_{i=1}^n collect(x, s_i)\]
where $n$ is the total number of sugars, $s_i$ is the $i$-th sugar. And the function $collect$ returns 1 if $s_i$ was collected by the agent $x$, otherwise it returns 0. Once a sugar is collected by a specific agent, no other agent can collect it and get rewarded again, also sugars do not re-spawn. The main loop of the algorithm is described in depth in \nameref{sec:NN_config}.\\
The visualization of the sugar placement in a maze is shown in Figure \ref{fig:sugars1}, where the hard map that was used in our experiments is depicted. It is important to highlight that the sugars are placed randomly and their density is the only hyperparameter.\\

\subsection{Pixel Novelty}
Naturally, we started thinking of ways to generalize Sugar Search in the games, where we do not have access to the position of the agent. A good example of this problem are Atari Games, where the only information we have is the screen. Therefore, agent localization and tracking are difficult. To overcome this problem, we propose Pixel Novelty, a generalization of our Sugar Search. Instead of rewarding the agent for collecting the sugar first, we will reward him for generating a new set of pixels (new screen content) first. Every time this occurs, the agent's fitness value is increased and the pixel compositions of the new screen are stored for future comparisons. If this or any other agent creates the same screen in the current generation, he is not rewarded again. Importantly, the archive of screens is reset after each generation, therefore in the next run, the same logic is applied. This idea promotes the agents to act in an unexpected, new, and never-before-experienced way, and it also solves the problem of agent localization and tracking. Eliminating this process is expected to reduce the computational time significantly.\\

\section{Experiments}
\subsection{Maze navigation task}
To prove our suggested approach - Sugar Search and to see how it compares to the objective-based search and Novelty Search, we decided to test these algorithms on the maze environments. These environments, proposed in \citep{b3}, are widely used due to the fact that they contain many dead ends and local optima. Algorithms that are optimizing the distance to the goal may be caught in a local minimum and thus perform worse than the divergent algorithms like ours.\\
\begin{figure}[h]
\centering
\begin{subfigure}[b]{0.23\textwidth}
   \includegraphics[width=0.8\linewidth]{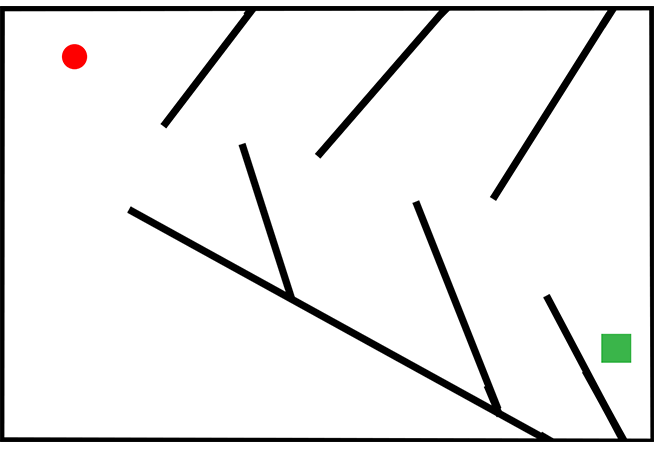}
   \caption{Medium map. A non-trivial wavy movement is required to reach the goal.}
   \label{fig:mazesMed}
\end{subfigure}
\hspace*{0.0005\textwidth}
\begin{subfigure}[b]{0.23\textwidth}
   \includegraphics[width=0.8\linewidth]{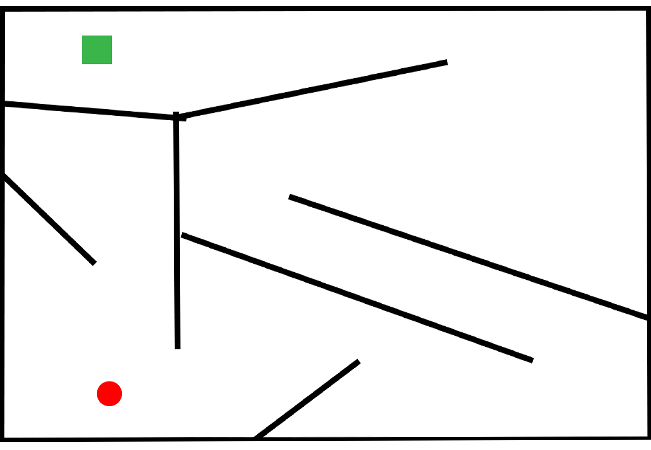}
   \caption{Hard map. Agents have to move further away from the goal in order to reach it.}
    \label{fig:mazesHard}
\end{subfigure}
\begin{subfigure}[b]{0.23\textwidth}
   \includegraphics[width=0.8\linewidth]{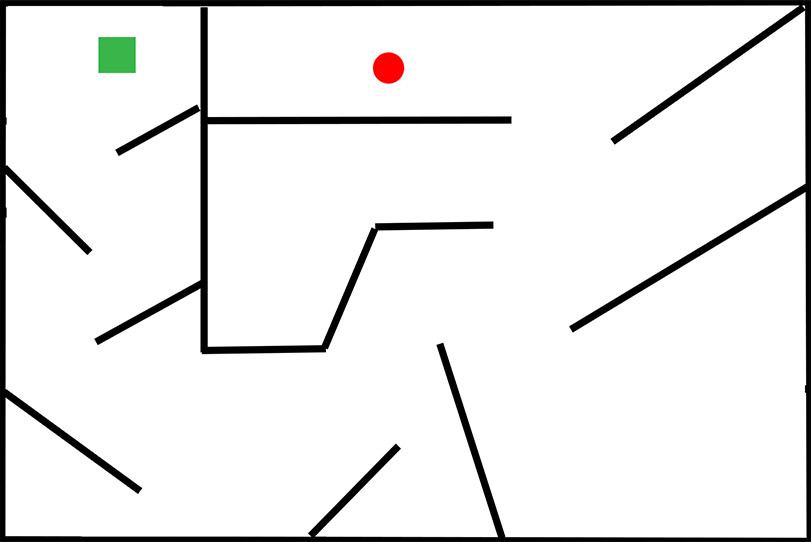}
   \caption{Super-hard map. It introduces more dead ends and a complex movement of the agent is needed to pass it successfully.}
    \label{fig:superhardmap}
\end{subfigure}
\caption{Medium, hard and super-hard map. The red dot is the starting point, the green dot is the goal, and the black lines are the obstacles.}
\label{fig:mazes}
\end{figure}

\begin{figure}[h!]
\centering
\includegraphics[scale=0.32]{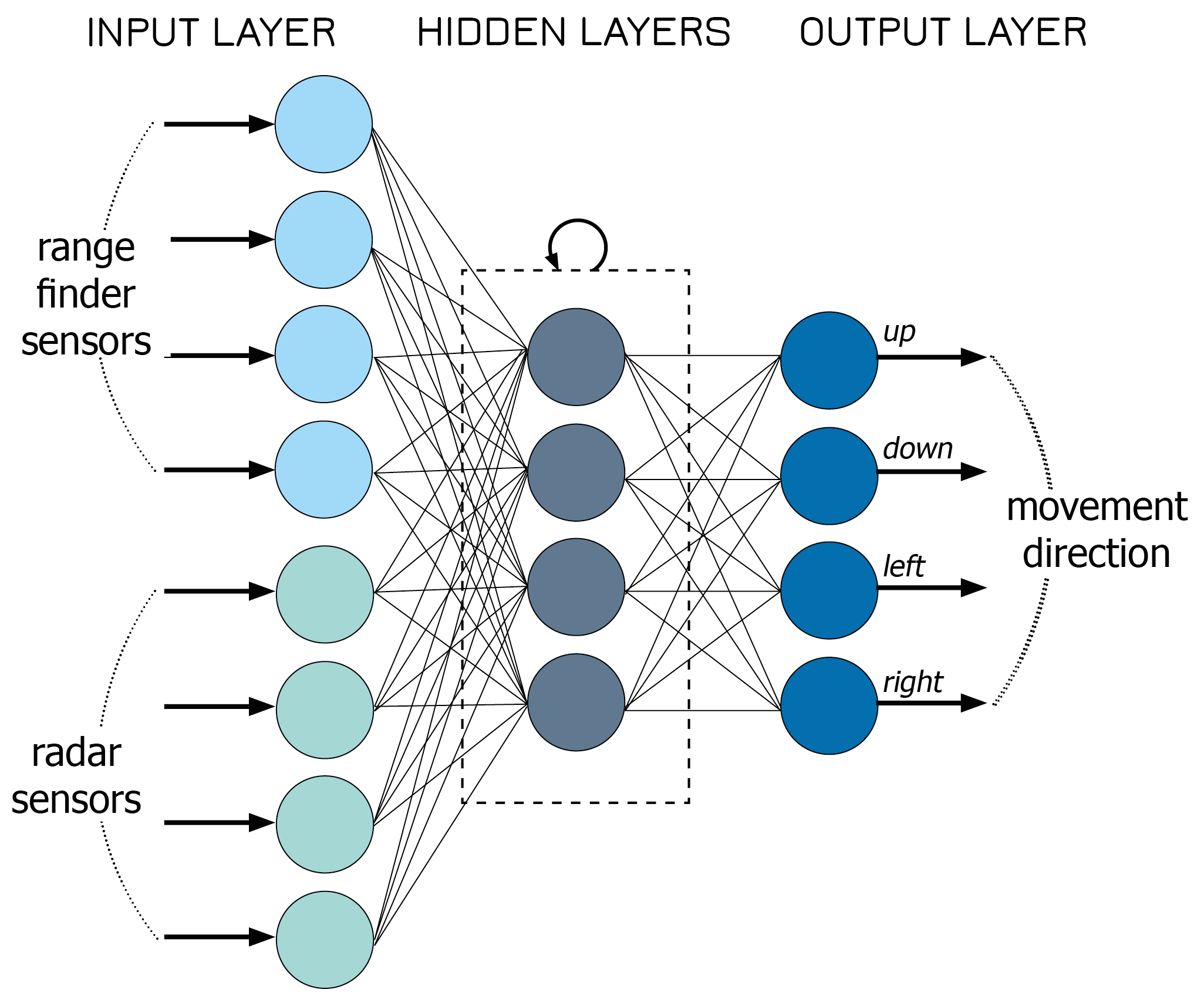}
\caption{Agent's neural network in the maze navigation task.}
\label{fig:agent1NN}
\end{figure}

The maze navigation task takes place in a set of 3 maze environments of increasing difficulty, which are shown in Figure \ref{fig:mazes}. The maps of these mazes were designed to correspond with the mazes from the original Novelty Search paper \citep{b3}. The medium map in Figure \ref{fig:mazesMed} has deceptive dead ends that can lead the agent to a local minimum. To overcome this problem, the agent has to develop a wavy movement to get through the obstacles to the goal. The second maze, the hard map, is more difficult because the agent has to move completely away from the goal in order to reach it, as we can observe in Figure \ref{fig:mazesHard}.\\
Finally, we have also constructed the super-hard maze shown in Figure \ref{fig:superhardmap}, which is a combination of the medium and the hard mazes. It contains deceptive ends and also requires a non-trivial movement to reach the goal.\\
For the maze navigation task, sugars were distributed randomly and uniformly over the environment with a density of 0.3 per pixel as listed in Table \ref{tab_app1}. In the fitness-based search, a distance to the goal was used to determine the fitness value of the agent. In the Novelty Search technique, 15-nearest neighbors were used.\\
All experiments were done under the same conditions and all the configuration specifics are listed in Table \ref{tab_app1}. The agent that gets close enough to the goal, within a fixed amount of time, is considered to be the solution.\\
Throughout the experiments, we use agents that are controlled by a neural network (NN) shown in Figure \ref{fig:agent1NN}. The goal of an agent is to develop such a NN that it can navigate him from the starting point to the endpoint of the maze before the timeout. Each agent is equipped with 8 sensors in total: 4 range finder sensors, which show the distance to the nearest obstacle on each side, and 4 radar sensors, which indicate the direction to the goal. The value of these sensors is normalized and used as the input to the neural network, which will evaluate whether the agent should move up, down, left or right. The specifics of the NN that was used throughout all the experiments are further described in \nameref{sec:NN_config}.\\

\subsection{Atari games task}
The maze navigation task is a well-suited environment to test our approach and compare it to other works such as Novelty Search. More advanced problems such as walking robots could show the same results.\\
Further, we decided to implement this idea and generalize it to the environment of Atari games. Atari games have been widely used to compare many machine learning algorithms, especially deep reinforcement learning algorithms \citep{dqn}. But due to the many implementations of Atari games, it is difficult to compare the results of different algorithms. That is why the OpenAI Gym \citep{open_ai} games were chosen at the beginning to avoid any differences in the underlying environment and to see if our ideas could compare with some of the famous algorithms \citep{dqn}\citep{ac}.\\
Later on, due to the difficulties with the agent localization and tracking in the OpenAI Gym \citep{open_ai}, we have decided to test our Sugar Search approach on the MinAtar \citep{minatar}. MinAtar is inspired by the Arcade Learning Environment \citep{ALE} but simplifies the games to make experiments with the environments more accessible and efficient. Currently, MinAtar provides analogues to 5 Atari games, which we have extended by Ms. Pacman and Montezuma's revenge implementation. The environments provide a state representation, where each of the $n$ channels corresponds to a game-specific object, such as the ball, the paddle, and the brick in the game Breakout. Thanks to this environment, we were able to track the agent and perform Sugar Search.\\
The goal of the agent is similar to the maze navigation task: to develop such NN that it can achieve the highest score possible. In the MinAtar environment, the agent's input consists of all the state representations of the objects. In the OpenAI gym, the input is the screen pixels. However, for computational reasons, the screen resolution was down-sampled 8 times to $(210/8)\times(160/8)$ pixels, and the RGB colors were converted into gray-scale. This alteration resulted in 520 input neurons used in the experiments, from the original 100,800.\\
All the necessary configuration parameters used in the Atari games task are included in Table \ref{tab_app1}. In the fitness-based search, a score was used to determine the fitness value of the agent. In the Novelty Search technique, 5-nearest neighbors were used. However, to reach even better performance, we had to modify the idea of our approach slightly by re-spawning a new sugar during the game. Because in some Atari games, e.g. Ms. Pacman, the agent has no finish point and its goal is to collect as many points as possible before getting killed. This requires traveling to places that have been explored in the past but may be beneficial to revisit in the present moment. To motivate the agent to behave in such a way, we decided to re-spawn a sugar every $l$ frame.\\

\subsection{Neural Network configuration}
\label{sec:NN_config}
\begin{table}[h!]
\caption{The experiment configuration for both the Maze navigation task and the Atari games of our Sugar Search with RNN.}
\begin{center}
\begin{tabularx}{\linewidth}{|X|p{1.5cm}|p{1.5cm}|}
\hline
\textbf{Parameters} & \textbf{\textit{Maze env.}} & \textbf{\textit{Atari env.}}\\
\hline
Population size& 250 & 75 \\
Max. generations& 1200 & 10000 \\
Time frame (max. frames) & 600 & 250\\
Sugars density per pixel & 0.3 & 0.3 \\
$l$ re-spawn time frame & - & 5  \\
Hidden layer& 1  & 1\\
Hidden neurons& 32 & 32\\
Activation function& RelU & RelU\\
Gaussian noise& $\mathcal{N}(0,\,0.1^{2})$ & $\mathcal{N}(0,\,0.1^{2})$\\
\hline
\end{tabularx}
\end{center}
\label{tab_app1}
\end{table}

Each agent has to develop a neural network, which will allow him to achieve the goal. The challenge is how to develop this neural network. In the original paper \citep{b3} NEAT, an evolutionary algorithm was used \citep{b5}. However, it has numerous hyperparameters which are difficult to optimize. Instead, we used Elman recurrent neural network \citep{elman} (RNN), which is a simple NN with a recurrent hidden layer. We used random Gaussian noise to mutate its weights.\\
In our algorithm, we randomly initialize weights to the RNN for each agent. When the time limit is reached, we evaluate 10\% of the best agents and keep them for the next generation. The rest of the population is modified - random Gaussian noise is added to its weights. After this mutation, the agents are added to the next generation. This approach is much simpler and results in fewer hyperparameters.\\
The benefits of our concept of the NN are obvious from the experiments on the medium map, where NEAT needed 19.8 generations on average to finish the maze, whereas the simple Elman neural network \citep{elman} with our evolutionary algorithm only needed 10.4 generations (these results were averaged from 50 runs). This proves that our approach is much better, and for these reasons we used it in all of our experiments. All the other parameters necessary for the reproduction of the experiments are listed in Table \ref{tab_app1}.\\

\section{Results}
\subsection{Maze navigation task}
We have evaluated the results from the least to the most complex map: the medium map, then the hard map, and finally the super-hard map in Figure \ref{fig:mazes}. All the results are shown in detail in Table \ref{tab1}. The level of success of the search techniques is measured by the average number of generations needed to reach the goal and the number of successfully finished runs out of 50 (meaning runs where we were able to reach the goal). Note that we use the same way to reference the maps as in the Novelty Search \citep{b3} paper to avoid confusion.\\
On the medium map, Novelty Search and our Sugar Search performed similarly, while the fitness-based search was approximately two times slower. All of these techniques were able to successfully finish all 50 out of 50 runs. The difference between divergent searches and fitness-based search is statistically significant ($p<0.01$; Student's t-test). 
On the hard map, where the deceptiveness of the environment is more compelling, a remarkable difference between all three examined algorithms can be distinguished. Sugar Search performed significantly better than Novelty Search in terms of the success rate and the number of generations needed, as shown in Table \ref{tab1}. The difference between Sugar Search and Novelty Search is statistically significant ($p<0.05$; Student's t-test). Both divergent approaches largely outperformed the fitness-based search, which was unable to reach the goal in any run. \\
Similar results were observed on the super-hard map. The fitness-based approach failed to reach the goal in any run again. The divergent approaches were able to solve this maze with a reasonable success rate: 14 for Novelty Search versus 24 for Sugar Search, and with an average number of generations needed: 345.3 for Novelty Search versus 280.4 for Sugar Search. This difference is not statistically significant ($p<0.05$; Student's t-test). A larger sample of runs on this map would be needed to confirm the significant difference of our Sugar Search outperforming Novelty Search, which can be observed in the results.

\setlength{\tabcolsep}{0.5em} 
{\renewcommand{\arraystretch}{1.2}
\begin{table}[h!]
\caption{Maze navigation task performance in the average number of generations needed to reach the goal and the number of successfully
finished runs out of 50 (meaning runs where we were able to
reach the goal). The standard deviation is in parentheses. Bold font indicates the best performance for each metric. All numbers are averaged on the successful runs only. Symbols $\uparrow$ ($\downarrow$) represent that the higher (lower) the better.}
\begin{center}
\begin{tabularx}{\linewidth}{ |p{1.3cm}|c|c|c|c|c|c| }
\hline
\textbf{Maze Results} & \multicolumn{3}{p{2.5cm}|}{\textbf{Generations $\downarrow$}} & \multicolumn{3}{p{2.5cm}|}{\textbf{Finished runs (out of 50) $\uparrow$}} \\
\cline{2-7}
 & \multicolumn{1}{p{0.7cm}|}{\textbf{\textit{med. map}}} & \multicolumn{1}{p{0.7cm}|}{\textbf{\textit{hard map}}} & \multicolumn{1}{X|}{\textbf{\textit{super-hard map}}}
 & \multicolumn{1}{p{0.7cm}|}{\textbf{\textit{med. map}}} & \multicolumn{1}{p{0.7cm}|}{\textbf{\textit{hard map}}} & \multicolumn{1}{X|}{\textbf{\textit{super-hard map}}} \\
\hline
\multirow{3}{1.3cm}{Fitness-based approach} & \multirow{2}{0.6cm}{\centering17.6 \scriptsize{(9.18)}} & - & - & 50 & 0 & 0 \\
&&&&&&\\
&&&&&&\\\hline
\multirow{2}{1.3cm}{Novelty approach} & \multirow{2}{0.6cm}{\centering\textbf{10.4} \scriptsize{(5.08)}} & \multirow{2}{0.7cm}{\centering237.7 \scriptsize{(172.12)}} & \multirow{2}{0.7cm}{\centering345.3 \scriptsize{(183.42)}} & 50 & 23 & 14 \\
&&&&&&\\\hline
\multirow{3}{1.3cm}{Sugar Search (ours)} & \multirow{2}{0.6cm}{\centering11.7 \scriptsize{(5.96)}} & \multirow{2}{0.7cm}{\centering\textbf{153.3} \scriptsize{(97.53)}} & \multirow{2}{0.7cm}{\centering\textbf{280.4} \scriptsize{(101.78)}} & 50 & \textbf{35} & \textbf{24} \\
&&&&&&\\
&&&&&&\\\hline
\end{tabularx}
\end{center}
\label{tab1}
\end{table}
}

\subsection{Reward density}\label{AA}
\begin{figure}[h!]
\begin{center}
\includegraphics[scale=0.52]{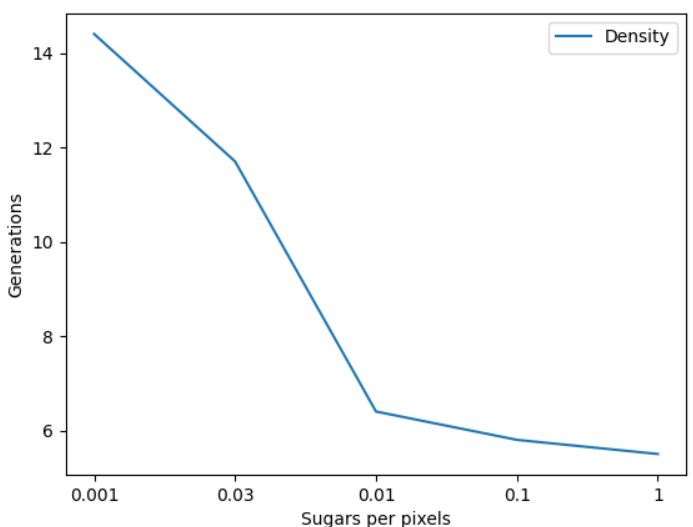}
\caption{Average number of generations
needed to reach the goal (\textit{the lower the better}) and the sugars per pixel density trade-off curve.}
\label{fig:sugarsdensity}
\end{center}
\end{figure}

Naturally, while working on the maze problem, an interesting question arose. How does the density of the reward distribution on the map affect the results? Our hypothesis was that having as many sugars as there are pixels on the map would produce the best results since theoretically a higher density of sugars should lead agents to the goal more efficiently.\\
From a series of experiments on the medium map with a varying sugar density parameter as shown in Figure \ref{fig:sugarsdensity}, we have proven this hypothesis valid. Placing sugar on every pixel produces the best results, and reducing the sugar density generally slows down the process of reaching the goal. Interestingly, performing a small-scale reduction of the reward density still produces comparable results. \\
These findings can be beneficial when applied to a similar problem defined in a continuous space, where the sugar placement cannot be realized per pixel for computational reasons. \\

\subsection{Combining Sugar Search with the fitness-based approach}
\label{combination}
We could observe that in a deceptive maze problem, the divergent search, especially our Sugar Search, completely outperforms the objective-driven search. But what if we combined these two approaches? \\
We wanted to explore this idea further and decided to design a combined approach, where the agent is rewarded not only for being novel (divergent search) but also for moving towards the goal (fitness-based search). This combined approach requires us to define the weights of each component that would produce the best results, meaning we need to find an optimal alpha.
\[p(x)=\alpha*distance + (1-\alpha)*sugars\]

\setlength{\tabcolsep}{0.5em} 
{\renewcommand{\arraystretch}{1.2}
\begin{table}[h!]
\caption{Maze navigation task with the weighted approach: performance in the average number of generations needed to reach the goal and the number of successfully
finished runs out of 50 (meaning runs where we were able to reach the goal). The standard deviation is in parentheses. Bold font indicates the best performance for each metric. All numbers are averaged on the successful runs only. Symbols $\uparrow$ ($\downarrow$) represent that the higher (lower) the better.}
\begin{center}
\begin{tabularx}{\linewidth}{|p{2.5cm}|c|c|c|c|}
\hline
\textbf{Maze Results} & \multicolumn{2}{p{2.4cm}|}{\textbf{Generations $\downarrow$}} & \multicolumn{2}{X|}{\textbf{Finished runs (out of 50) $\uparrow$}} \\
\cline{2-5} 
 & \multicolumn{1}{p{1.2cm}|}{\textbf{\textit{medium map}}} & \multicolumn{1}{p{0.8cm}|}{\textbf{\textit{hard map}}}
 & \multicolumn{1}{p{1.2cm}|}{\textbf{\textit{medium map}}} & \multicolumn{1}{p{0.8cm}|}{\textbf{\textit{hard map}}} \\
\hline
Fitness-based & \multirow{2}{0.7cm}{\centering17.6 \scriptsize{(9.18)}} & - & 50 & 0 \\
approach &&&& \\
\hline
Novelty approach & \multirow{2}{0.7cm}{\centering\textbf{10.4} \scriptsize{(5.08)}} & \multirow{2}{0.7cm}{\centering237.7 \scriptsize{(172.12)}} & 50 & 23 \\
approach &&&& \\
\hline
Sugars & \multirow{2}{0.7cm}{\centering11.7 \scriptsize{(5.96)}} & \multirow{2}{0.7cm}{\centering153.3 \scriptsize{(97.53)}} & 50 & \textbf{35} \\
approach &&&& \\
\hline
Weighted & \multirow{2}{0.7cm}{\centering8.0 \scriptsize{(7.97)}} & \multirow{2}{0.7cm}{\centering148.7 \scriptsize{(104.25)}} & 50 & 32 \\
with alpha 1/4 &&&& \\
\hline
Weighted & \multirow{2}{0.7cm}{\centering\textbf{5.4} \scriptsize{(4.09)}} & \multirow{2}{0.7cm}{\centering\textbf{122.4} \scriptsize{(86.55)}} & 50 & 30 \\
with alpha 2/4 &&&& \\
\hline
Weighted & \multirow{2}{0.7cm}{\centering6.7 \scriptsize{(5.59)}} & \multirow{2}{0.7cm}{\centering139.1 \scriptsize{(90.87)}} & 50 & 26 \\
with alpha 3/4 &&&& \\
\hline
\end{tabularx}
\end{center}
\label{tab4}
\end{table}
}

In order to find an optimal alpha and observe how this weighted approach stands in comparison with the original Sugar Search and other techniques, we have executed a series of experiments on the medium and the hard maps. To reduce the computation time, as a proof of concept, we have excluded the super-hard map from the following experiments. Based on the results of the basic Sugar Search algorithm, we can expect to observe similar trends on the super-hard map as the hard map.\\
From the results shown in Table \ref{tab4}, we can conclude that the combined approach leads to improved performance, specifically in the average number of generations needed to reach the goal. On the other hand, an improvement is not observable in the agent's success rate. It was 29.3 on average on the hard map, which is slightly worse than Sugar Search alone. \\

\subsection{Agents without sensors}

The idea behind this section is to prove that the agents are really processing the information from their sensors to avoid hitting an obstacle and that they are not simply memorizing a sequence of optimal actions instead. This problem could arise when we take into consideration the amount of memorization and the overall generalization in terms of the agents' performance in the mazes.\\
\begin{figure}[h!]
\centering
\begin{subfigure}[b]{0.45\textwidth}
   \includegraphics[width=0.8\linewidth]{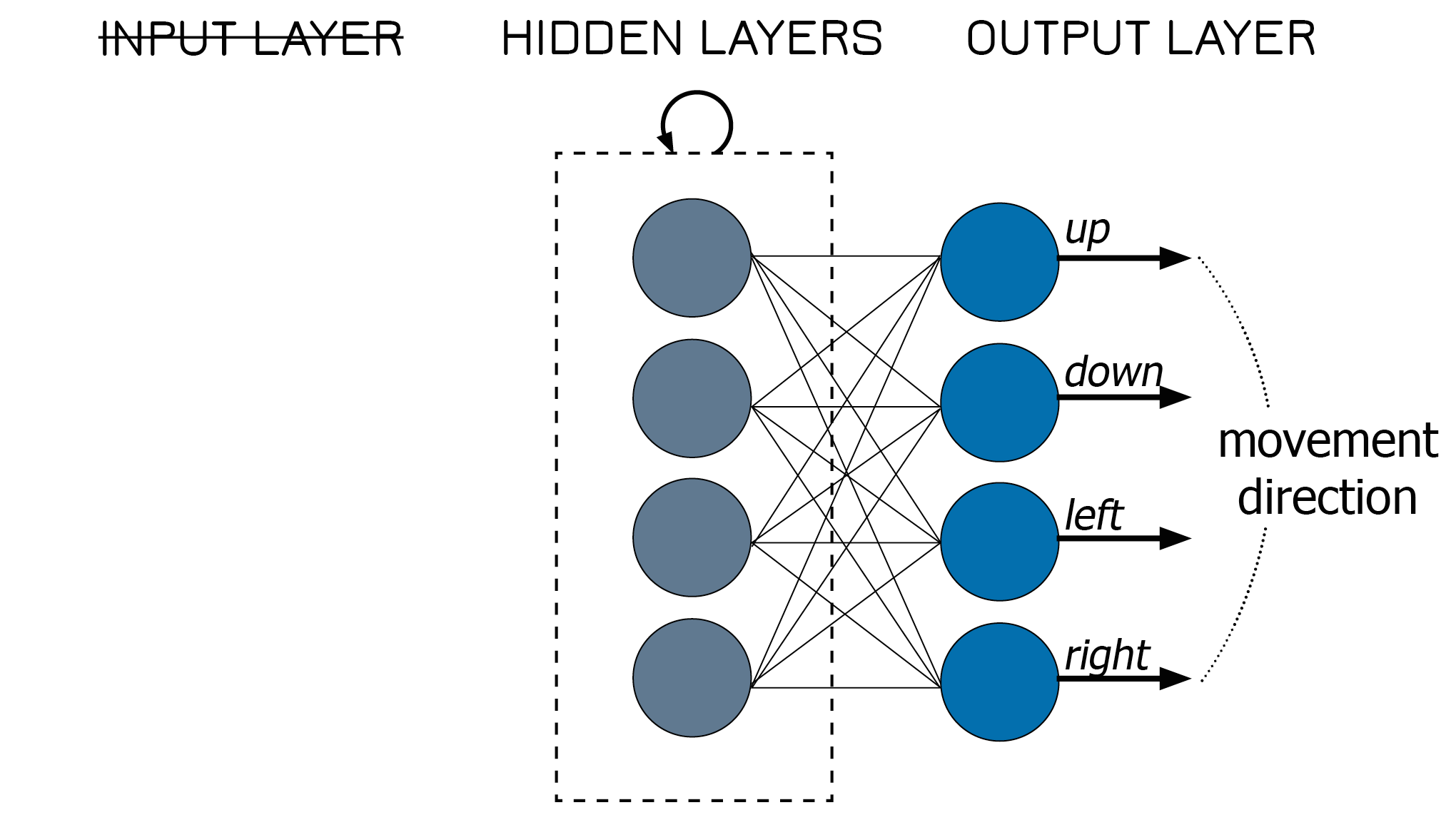}
   \caption{A blind agent relies solely on the memory in the hidden layer.}
   \label{fig:agentBlind}
   \vspace{0.4cm}
\end{subfigure}
\begin{subfigure}[b]{0.45\textwidth}
   \includegraphics[width=0.8\linewidth]{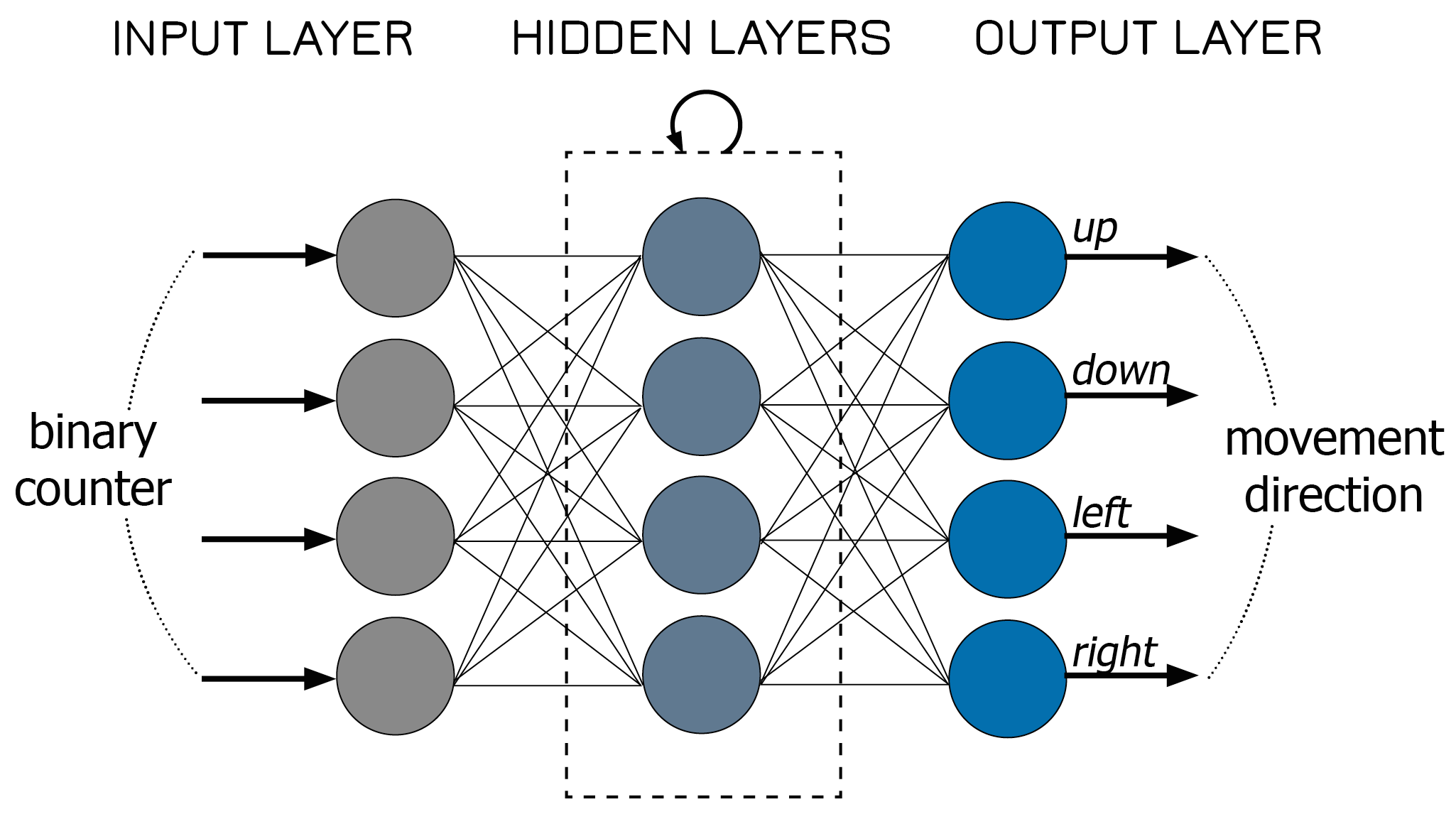}
   \caption{A modified blind agent with a simple binary counter on the input.}
   \label{fig:agentBinCounter}
\end{subfigure}
\caption{Modified agents without the sensors and their NN.}
\label{fig:modifiedAgents}
\end{figure}
To prove this, we have decided to experiment with a blind agent. The original agents were blinded - meaning they had all their inputs removed and had to rely solely on their memory in the hidden layer, as shown in Figure \ref{fig:agentBlind}. Later, we have modified this concept further in order to elevate the memory and created a modified blind agent that had a simple binary counter as its input as shown in Figure \ref{fig:agentBinCounter}. This binary counter is a timer, which helps the agent memorize sequences of actions. \\
From Table \ref{tab2} we can see that the blind agents can navigate throughout the map successfully without having any information about the environment, simply by memorizing a sequence of actions. However, having information from the sensors about the nearest obstacle and the location of the goal significantly accelerates the process of reaching the goal. It is also important to note that the blind agents are not adaptable and will fail completely when placed in a slightly different starting point or if the task is modified.\\

\setlength{\tabcolsep}{0.5em} 
{\renewcommand{\arraystretch}{1.2}
\begin{table}[h]
\caption{Maze navigation task for agents without sensors: performance in the average number of generations needed to reach the goal. The standard deviation is in parentheses. Bold font indicates the best performance for each metric. Symbols $\uparrow$ ($\downarrow$) represent that the higher (lower) the better.}
\begin{center}
\begin{tabular}{|p{1.83cm}|c|p{1.7cm}|p{1.88cm}|}
\hline
\cline{2-3} 
\textbf{Maze} & \textbf{\textit{Sensors $\downarrow$}}& \textbf{\textit{Binary}} & \textbf{\textit{No input $\downarrow$}}\\
\textbf{Results} && \textbf{\textit{counter $\downarrow$}} &\\
\hline
medium map& \textbf{11.7} \scriptsize{(5.96)} & 37.8 \scriptsize{(50.88)} & 96.1 \scriptsize{(36.91)} \\
hard map& \textbf{153.3} \scriptsize{(97.53)} & 597.4 \scriptsize{(354.65)} & 1154.7 \scriptsize{(592.78)} \\
\hline
\end{tabular}
\end{center}
\label{tab2}
\end{table}
}

\subsection{MinAtar results}
In this environment, we ran experiments on 7 different Atari games to test and compare the fitness-based approach, Sugar Search, and its generalization - Pixel Novelty to other techniques like Novelty Search \citep{b3} or DQN \citep{dqn}. The results in Table \ref{tab5} show that Sugar Search does not always lead to a higher score compared to the fitness-based approach, but it performs comparably well with Novelty Search consistently across different games, as shown in Table \ref{tab5}. The scores achieved with Pixel Novelty are also competitive or comparable to other approaches, meaning that Pixel Novelty is a valid generalization of Sugar Search.\\
From the results in Table \ref{tab5}, we can conclude that the achieved scores differ based on the characteristics of each game, where some approaches seem to be more fitted than others. Generally speaking, we can observe a pattern in Atari games with no distinctive local optima, such as Breakout or Space Invaders, where a simple fitness-based search outperforms both our Sugar Search and Novelty Search. Interestingly, Pixel Novelty also performs rather well in these games (it even achieved the highest score in Breakout).\\
Regarding Sugar Search, it outperformed both the fitness-based search and Novelty Search by far in games like Freeway or Ms. Pacman and was even able to compete with algorithms like DQN \citep{dqn} despite its simplicity. This demonstrates that our search technique is very powerful and able to produce competitive results.\\

\setlength{\tabcolsep}{0.5em} 
{\renewcommand{\arraystretch}{1.2}
\begin{table}[h]
\caption{MinAtar games performance in the average number of the achieved game score. Bold font indicates the best performance for each metric. All numbers are averaged from 100 generations/episodes. Symbols $\uparrow$ ($\downarrow$) represent that the higher (lower) the better.}
\begin{center}
\begin{tabularx}{\linewidth}{ |X|ccccc| }
\hline
\textbf{Game} & \multicolumn{1}{p{0.95cm}|}{\textbf{\textit{Fitness-based Search $\uparrow$}}} & \multicolumn{1}{p{0.85cm}|}{\textbf{\textit{Sugar Search (ours) $\uparrow$}}} & \multicolumn{1}{p{1cm}|}{\textbf{\textit{Novelty Search $\uparrow$}}} &
\multicolumn{1}{p{0.9cm}|}{\textbf{\textit{Pixel Novelty (ours) $\uparrow$}}}&
\multicolumn{1}{p{0.5cm}|}{\textbf{\textit{DQN $\uparrow$ }}}
\\ \hline
Seaquest & 8 & 7 & 6 & 5&\textbf{20}\\\cline{1-1}
Breakout & 13 & 5 & 5 & \textbf{21}& 9\\ \cline{1-1}
Asterix & 12 & 10 & 14 & 10& \textbf{17} \\\cline{1-1}
Freeway & 16 & \textbf{53} & 49 & 28 & 50 \\\cline{1-1}
S. Invaders & 43 & 10 & 14 & 21 & \textbf{45} \\\cline{1-1}
MsPacman & 1897 & \textbf{3580} & 1962 & 2480 & 1099 \\\cline{1-1}
Montezuma's revenge & 0 & 41 & 22& 22 & \textbf{46}\\\hline
\end{tabularx}
\end{center}
\label{tab5}
\end{table}
}

\subsection{OpenAI Gym results}
In the OpenAI Gym environment, experiments on 12 Atari games in total were performed with both Pixel Novelty and the fitness-based approach. The specific games to be included in the experiments were selected to represent equal subsets of games, half of which contains multiple obvious local minima (e.g. Montezuma's revenge) and is expected to show the advantages of our approach. The results of the experiments are presented in Table \ref{tab6} in comparison with the fitness-based approach and reinforcement learning algorithms like DQN \citep{dqn} and A3C algorithm \citep{a3c}.\\

\setlength{\tabcolsep}{0.5em} 
{\renewcommand{\arraystretch}{1.2}
\begin{table}[h]
\caption{OpenAI Gym games performance in the average number of the achieved game score. Bold font indicates the best performance for each metric. All numbers are averaged from 100 generations/episodes. Symbols $\uparrow$ ($\downarrow$) represent that the higher (lower) the better.}
\begin{center}
\begin{tabularx}{\linewidth}{|X|cccc|}
\hline
\textbf{Game} & \multicolumn{1}{p{1.2cm}|}{\textbf{\textit{Fitness-based $\uparrow$}}} & \multicolumn{1}{p{1.2cm}|}{\textbf{\textit{Pixel Novelty (ours) $\uparrow$}}} & \multicolumn{1}{p{1.2cm}|}{\textbf{\textit{DQN $\uparrow$ }}} &
\multicolumn{1}{p{1.2cm}|}{\textbf{\textit{A3C FF $\uparrow$ }}} \\ \hline
BeamRider & 1188.6 & 900.4 & 8672.4 & \textbf{13235.9} \\ \cline{1-1}
Breakout & 15.4 & 24.1 & 303.9 & \textbf{551.6} \\ \cline{1-1}
Enduro & 86.9 & 52.2 & \textbf{475.7} & 82.2 \\ \cline{1-1}
Pong & 21.0 & \textbf{24.3} & 16.2 & 11.4 \\ \cline{1-1}
Seaquest & 780.4 & 480.3& \textbf{2793.9} & 2300.2\\ \cline{1-1}
S. Invaders & 1075.7 & 1205.3 & 1449.7 & \textbf{2214.7} \\ \cline{1-1}
Riverraid & 2390.5 & 1810.7 & 4065.3 & \textbf{10001.2} \\ \cline{1-1}
Freeway & \textbf{30.0} & 0.4  & 25.8 & 0.1 \\ \cline{1-1}
Gravitar & \textbf{2350.4} & 700.6 & 216.5 & 269.5 \\ \cline{1-1}
Zaxxon & \textbf{6240.1} & 5400.4 & 831.0 & 2659.0 \\ \cline{1-1}
Venture & \textbf{440.5} & 200.3 & 54.0 & 19.0 \\ \cline{1-1}
Montezuma's revenge& 0.0 & 10.0 & 50.0 & \textbf{53.0} \\\hline
\end{tabularx}
\end{center}
\label{tab6}
\end{table}
}
In the light of the results, rewarding the agents solely for generating new pixels in Pixel Novelty leads to the same or better results as the score-driven reward system in the fitness-based approach. These overall positive results are the outcome of the agent's behavior in Pixel Novelty: as the agent tries to generate a new unseen screen content, he conquers obstacles on the way and by doing so also achieves a reasonable score.\\
Additionally, when compared to DQN and A3C algorithms, we are able to achieve a higher score in 5 out of 12 games. Furthermore, it is important to highlight that our results were achieved with a significantly greater efficiency, since we only needed 20 million frames and 1 CPU for 24 hours, whereas the results from DQN and A3C were trained on 320 million frames, which is 16 times more.\\
We have proven that our simple evolutionary technique is a scalable competitive alternative to the reinforcement-learning algorithms \citep{es_vs_rl} and that it can achieve similar results even with a divergent-driven approach and with a largely improved efficiency.\\

\section{Conclusion}
The idea of placing rewards within the environment to precede the problem of local minima performs surprisingly well in the maze experiments, especially taking into consideration that this concept completely disregards the objective (score). As the experiments have shown, our technique outperformed the popular Novelty Search \citep{b3} in both the success rate and the solution speed, furthermore, it is also conceptually simpler and easier to implement. Following this proof of concept of our technique, we have also explored the effect of multiple local minima and different reward densities on the search efficiency. The results indicate that the sparse rewards model could be applied to problems in a continuous or a higher-dimensional space.\\
We have subsequently generalized our search technique and studied how it performs in a more general and popular set of tasks - Atari games. Our technique performed comparably well with a basic fitness-based search and some of the most common reinforcement learning algorithms (DQN \citep{dqn}, A3C \citep{a3c}). We consider it to be a great achievement considering that our results were computed on much fewer frames than DQN or A3C.\\
In summary, we believe that it is rather the environment that formulates, motivates, and rewards individuals for being different and novel as we can observe in nature. Also supported by the fact that the concept has been proven valid on multiple experiments in this paper. This suggests that new ways of defining novelty, which avoid any form of supervision, can generally be successful and could be worth exploring in the future.

\section{Acknowledgements}
We would like to thank GoodAI for the research funding and valuable advice. We would also like to express appreciation to Daniela Hradilova for the useful discussions and suggestions. Our work is part of the RICAIP project that has received funding from the European Union’s Horizon 2020 research and innovation programme under grant agreement No. 857306.

\footnotesize
\bibliographystyle{apalike}
\bibliography{main} 

\end{document}